\documentclass{article}
\usepackage{spconf,amssymb,amsmath,graphicx,booktabs}
\usepackage{multirow}
\usepackage{makecell}
\usepackage{url}
\usepackage{hyperref}

\usepackage{color}
\usepackage[utf8]{inputenc} 

\newcommand{\espneturl}{\url{https://github.com/espnet/espnet}}

\title{A Comparative Study on Transformer vs RNN in Speech Applications}
\name{\it Shigeki Karita$^1$,\\
(Alphabetical Order)
\it Nanxin Chen$^3$, Tomoki Hayashi$^{5,6}$, Takaaki Hori$^7$, Hirofumi Inaguma$^8$, Ziyan Jiang$^3$, \\
\it Masao Someki$^5$, Nelson Enrique Yalta Soplin$^2$, Ryuichi Yamamoto$^4$, Xiaofei Wang$^3$, Shinji Watanabe$^3$, \\
\it Takenori Yoshimura$^{5,6}$, Wangyou Zhang$^9$}
\address{
  $^1$NTT Communication Science Laboratories, $^2$Waseda University, $^3$Johns Hopkins University,\\ 
  $^4$LINE Corporation, $^5$Nagoya University, $^6$Human Dataware Lab. Co., Ltd., \\
  $^7$Mitsubishi Electric Research Laboratories, $^8$Kyoto University,
  $^9$Shanghai Jiao Tong University}

\usepackage[
backend=biber,
style=ieee,
doi=false,isbn=false,url=false,eprint=false
]{biblatex} 

\defbibheading{bibliography}[\refname]{}
\DeclareSourcemap{
	\maps[datatype=bibtex, overwrite=true]{
		\map{
			\step[fieldsource=booktitle,
			match=\regexp{.*Interspeech.*},
			replace={Proc. Interspeech}]
			\step[fieldsource=journal,
			match=\regexp{.*INTERSPEECH.*},
			replace={Proc. Interspeech}]
			\step[fieldsource=booktitle,
			match=\regexp{.*ICASSP.*},
			replace={ICASSP}]
			\step[fieldsource=booktitle,
			match=\regexp{.*International Conference on Acoustics, Speech and Signal Processing.*},
			replace={ICASSP}]
			\step[fieldsource=publisher,
			match=\regexp{.+},
			replace={{}}]
			\step[fieldsource=month,
			match=\regexp{.+},
			replace={{}}]
			\step[fieldsource=location,
			match=\regexp{.+},
			replace={{}}]
			\step[fieldsource=address,
			match=\regexp{.+},
			replace={{}}]
		}
	}
}

\newcommand{\tomokiedit}[1]{{#1}}

\begin{document}
%
\maketitle

\ninept

\abovedisplayskip=5pt
\belowdisplayskip=5pt

\begin{abstract}
Sequence-to-sequence models have been widely used in end-to-end speech processing, for example, automatic speech recognition (ASR), speech translation (ST), and text-to-speech (TTS). 
This paper focuses on an emergent sequence-to-sequence model called Transformer, which achieves state-of-the-art performance in neural machine translation and other natural language processing applications.
We undertook intensive studies in which we experimentally compared and analyzed Transformer and conventional recurrent neural networks (RNN) in a total of 15 ASR, one multilingual ASR, one ST, and two TTS benchmarks. 
Our experiments revealed various training tips and significant performance benefits obtained with Transformer for each task including the surprising superiority of Transformer in 13/15 ASR benchmarks in comparison with RNN.
We are preparing to release Kaldi-style reproducible recipes using open source and publicly available datasets for all the ASR, ST, and TTS tasks for the community to succeed our exciting outcomes.

\end{abstract}
\section{Introduction}
\label{sec:introduction}

Transformer is a sequence-to-sequence (S2S) architecture originally proposed for neural machine translation (NMT)~\cite{VaswaniNIPS2017_7181} that rapidly replaces recurrent neural networks (RNN) in natural language processing tasks. 
This paper provides intensive comparisons of its performance with that of RNN for speech applications; automatic speech recognition (ASR), speech translation (ST), and text-to-speech (TTS).

One of the major difficulties when applying Transformer to speech applications is that it requires more complex configurations (e.g., optimizer, network structure, data augmentation) than the conventional RNN based models.
Our goal is to share our knowledge on the use of Transformer in speech tasks so that the community can fully succeed our exciting outcomes with reproducible open source tools and recipes.

Currently, existing Transformer-based speech applications~\cite{speech-transformer,CrossVila2018,li2019close} still lack an open source toolkit and reproducible experiments while previous studies in NMT~\cite{ott-etal-2018-scaling,tensor2tensor-W18-1819} provide them.
Therefore, we work on an open community-driven project for end-to-end speech applications using both Transformer and RNN by following the success of \textit{Kaldi} for hidden Markov model (HMM)-based ASR~\cite{kaldi}.
Specifically, our experiments provide practical guides for tuning Transformer in speech tasks to achieve state-of-the-art results.

In our speech application experiments, we investigate several aspects of Transformer and RNN-based systems. 
For example, we measure the word/character/regression error from the ground truth, training curve, and scalability for multiple GPUs.

The contributions of this work are:
{
\setlength{\leftmargini}{15pt}  
\begin{itemize}
	\setlength{\itemsep}{1pt}      
	\setlength{\parskip}{0pt}      
	\setlength{\itemindent}{0pt}   
	\setlength{\labelsep}{4pt}     
    \item We conduct a larges-scale comparative study on Transformer and RNN with significant performance gains especially for the ASR related tasks.
    \item We explain our training tips for Transformer in speech applications: ASR, TTS and ST. 
    \item  We provide reproducible end-to-end recipes and models pretrained on a large number of publicly available datasets in our open source toolkit \textit{ESPnet}~\cite{espnet}\footnote{\espneturl}.
\end{itemize}
}

\subsubsection*{Related studies}

As Transformer was originally proposed as an NMT system~\cite{VaswaniNIPS2017_7181}, it has been widely studied on NMT tasks including hyperparameter search~\cite{DBLP:journals/pbml/PopelB18}, parallelism implementation~\cite{ott-etal-2018-scaling} and in comparison with RNN~\cite{lakew-etal-2018-comparison}.
On the other hand, speech processing tasks have just provided their preliminary results in ASR~\cite{speech-transformer,Zhou2018}, ST~\cite{CrossVila2018} and TTS~\cite{li2019close}.
Therefore, this paper aims to gather the previous basic research and to explore wider topics (e.g., accuracy, speed, training tips) in our experiments.


\section{Sequence-to-sequence RNN}
\label{sec:s2s}

\subsection{Unified formulation for S2S}

\begin{figure}
    \centering
    \vspace{-2mm}
    \includegraphics[width=0.8\columnwidth]{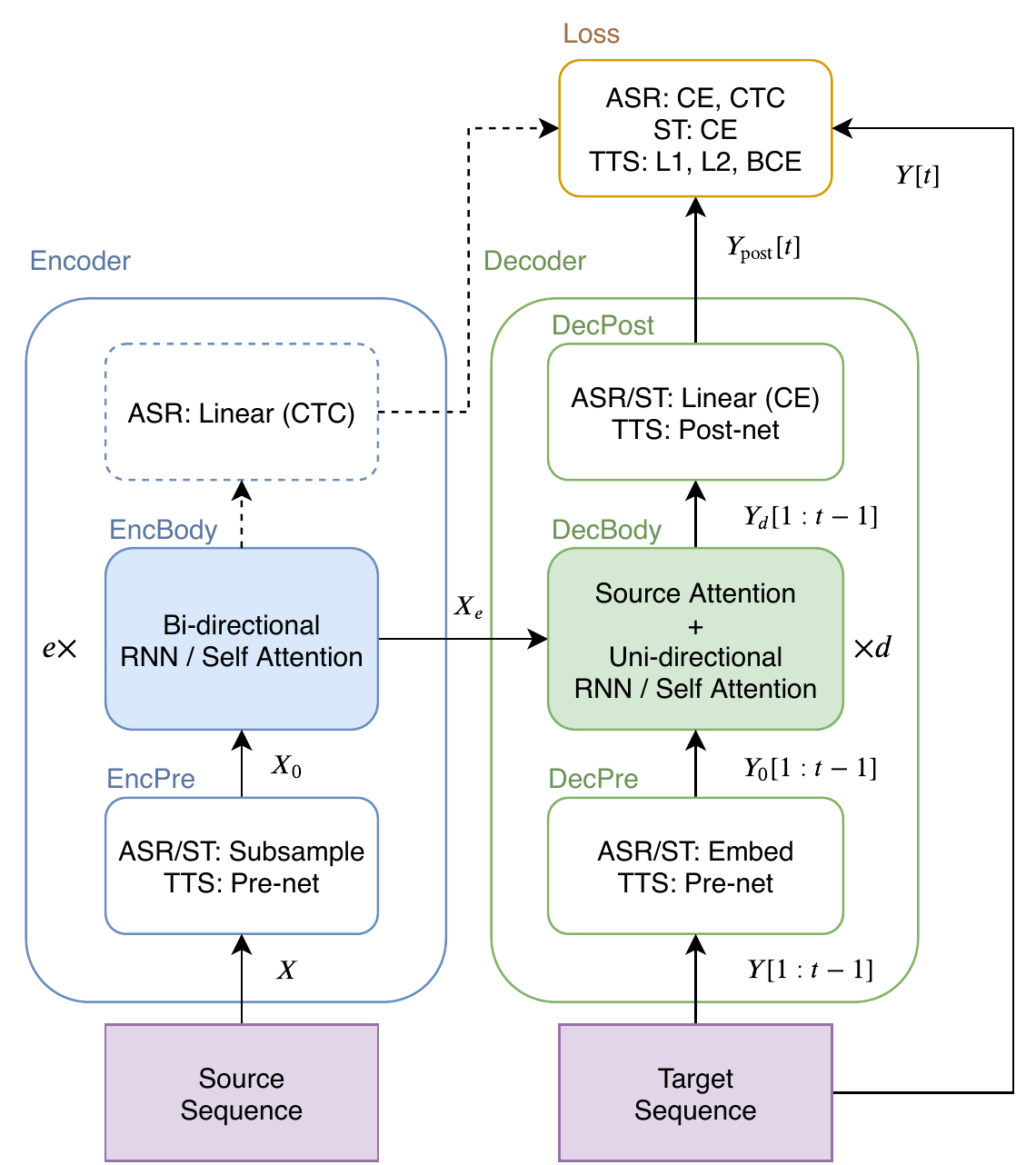}
    \vspace{-2mm}
    \caption{Sequence-to-sequence architecture in speech applications.}
    \vspace{-4mm}
    \label{fig:s2s}
\end{figure}

S2S is a variant of neural networks that learns to transform a source sequence $X$ to a target sequence $Y$~\cite{s2s_NIPS2014_5346}.
In Fig. \ref{fig:s2s}, we illustrate a common S2S structure for ASR, TTS and ST tasks.
S2S consists of two neural networks: an encoder
\begin{align}
\label{eq:encpre}
    X_0 &= \textrm{EncPre}(X), \\
\label{eq:encbody}
    X_e &= \textrm{EncBody}(X_0),
\end{align}
and a decoder
\begin{align}
\label{eq:decpre}
    Y_0[1:t-1] &= \textrm{DecPre}(Y[1:t-1]), \\
\label{eq:decbody}
    Y_d[t] &= \textrm{DecBody}(X_e, Y_0[1:t-1]), \\
\label{eq:post}
    Y_\textrm{post}[1:t] &= \textrm{DecPost}(Y_d[1:t]),
\end{align}
where $X$ is the source sequence (e.g., a sequence of speech features (for ASR and ST) or characters (for TTS)), $e$ is the number of  layers in $\textrm{EncBody}$, $d$ is the number of  layers in $\textrm{DecBody}$, $t$ is a target frame index, and all the functions in the above equations are implemented by neural networks. 
For the decoder input $Y[1:t-1]$, we use a ground-truth prefix in the training stage, while we use a generated prefix in the decoding stage.
During training, the S2S model learns to minimize the scalar loss value
\begin{align}
    L &= \textrm{Loss}(Y_\textrm{post}, Y)
\end{align}
between the generated sequence $Y_\textrm{post}$ and the target sequence $Y$.

The remainder of this section describes RNN-based universal modules: ``EncBody'' and ``DecBody''. We regard  ``EncPre'', ``DecPre'', ``DecPost'' and ``Loss'' as task-specific modules and we describe them in the later sections.

\subsection{RNN encoder}
\label{sec:rnn_enc}
$\textrm{EncBody}(\cdot)$ in Eq.~\eqref{eq:encbody} transforms a source sequence $X_0$ into an intermediate sequence $X_e$. Existing RNN-based $\textrm{EncBody}(\cdot)$ implementations~\cite{Bahdanau15,Chan2016,DBLP:conf/icassp/ShenPWSJYCZWRSA18} typically adopt a bi-directional long short-term memory (BLSTM) that can perform such an operation thanks to its recurrent connection. For ASR, an encoded sequence $X_e$ can also be used for source-level frame-wise prediction using connectionist temporal classification (CTC)~\cite{ctc-DBLP:conf/icml/GravesFGS06} for joint training and decoding~\cite{hori2018end}.

\subsection{RNN decoder}
\label{sec:rnn_dec}
$\textrm{DecBody}(\cdot)$ in Eq.~\eqref{eq:decbody} generates a next target frame with the encoded sequence $X_e$ and the prefix of target prefix $Y_0[1:t-1]$.
For sequence generation, the decoder is mostly unidirectional. For example, uni-directional LSTM with an attention mechanism~\cite{Bahdanau15} is often used in RNN-based $\textrm{DecBody}(\cdot)$ implementations. That attention mechanism emits source frame-wise weights to sum the encoded source frames $X_e$ as a target frame-wise vector to be transformed with the prefix $Y_0[0:t-1]$. We refer to this type of attention as ``encoder-decoder attention''.

\section{Transformer}
\label{sec:transformer}

Transformer learns sequential information via a self-attention mechanism instead of the recurrent connection employed in RNN.
This section describes the self-attention based modules in Transformer in detail.

\subsection{Multi-head attention}

Transformer consists of multiple dot-attention layers~\cite{luong-dot-att-D15-1166}:
\begin{align}
\label{eq:attn}
\mathrm{att}(X^\text{q}, X^\text{k}, X^\text{v}) = \mathrm{softmax}\left(\frac{X^\text{q} X^{\text{k} \top}}{\sqrt{d^\mathrm{att}}}\right) X^\text{v},
\end{align}
where $X^\text{k}, X^\text{v} \in \mathbb{R}^{n^\text{k} \times d^\mathrm{att}}$ and $X^\text{q} \in \mathbb{R}^{n^\text{q} \times d^\mathrm{att}}$ are inputs for this attention layer, $d^\mathrm{att}$ is the number of feature dimensions, $n^\text{q}$ is the length of $X^\text{q}$, and $n^\text{k}$ is the length of $X^\text{k}$ and $X^\text{v}$. We refer to $X^\text{q} X^{\text{k} \top}$ as the ``attention matrix''.
Vaswani et al.~\cite{VaswaniNIPS2017_7181} considered these inputs $X^\text{q}, X^\text{k}$ and $X^\text{v}$ to be a query and a set of key-value pairs, respectively.

In addition, to allow the model to deal with multiple attentions in parallel, Vaswani et al.~\cite{VaswaniNIPS2017_7181} extended this attention layer in Eq. (\ref{eq:attn}) to multi-head attention (MHA):
\begin{align}
\mathrm{MHA}(Q, K, V) &= [H_1, H_2, \dots, H_{d^\mathrm{head}}] W^\mathrm{head},\\
H_h &= \mathrm{att}(Q W^\text{q}_h, K W^\text{k}_h, V W^\text{v}_h),
\end{align}
where $K, V \in \mathbb{R}^{n^\text{k} \times d^\mathrm{att}}$ and $Q \in \mathbb{R}^{n^\text{q} \times d^\mathrm{att}}$ are inputs for this MHA layer, $H_h \in \mathbb{R}^{n^\text{q} \times d^\mathrm{att}}$ is the $h$-th attention layer output ($h = 1, \dots, d^\mathrm{head}$),  $W^\text{q}_h, W^\text{k}_h, W^\text{v}_h \in \mathbb{R}^{d^\mathrm{att} \times d^\mathrm{att}} $ and $ W^\mathrm{head} \in \mathbb{R}^{d^\mathrm{att} d^\mathrm{head} \times d^\mathrm{att}}$ are learnable weight matrices and $d^\mathrm{head}$ is the number of attentions in this layer.

\subsection{Self-attention encoder}
We define Transformer-based $\textrm{EncBody}(\cdot)$ used for Eq.~\eqref{eq:encbody} unlike the RNN encoder in Section~\ref{sec:rnn_enc} as follows:
\begin{align}
\nonumber
    \label{eq:self-attn}
    X_i' &= X_i + \mathrm{MHA}_i(X_i, X_i, X_i), \\
    X_{i+1} &= X_i' + \mathrm{FF}_i(X_i'),
\end{align}
where $i = 0, \dots, e - 1$ is the index of encoder layers,
and $\mathrm{FF}_i$ is the $i$-th two-layer feedforward network:
\begin{align}
    \mathrm{FF}(X[t]) = \mathrm{ReLU}(X[t] W^\text{ff}_1 + b^\text{ff}_1) W^\text{ff}_2 + b^\text{ff}_2,
\end{align}
where $X[t] \in \mathbb{R}^{d^\mathrm{att}}$ is the $t$-th frame of the input sequence $X$, $W^\text{ff}_1 \in \mathbb{R}^{d^\mathrm{att} \times d^\mathrm{ff}}, W^\text{ff}_2 \in \mathbb{R}^{d^\mathrm{ff} \times d^\mathrm{att}}$ are learnable weight matrices, and $b^\text{ff}_1 \in \mathbb{R}^{d^\mathrm{ff}}, b^\text{ff}_2 \in \mathbb{R}^{d^\mathrm{att}}$ are learnable bias vectors.
We refer to $\mathrm{MHA}_i(X_i, X_i, X_i)$ in Eq. (\ref{eq:self-attn}) as ``self attention''.

\subsection{Self-attention decoder}

Transformer-based $\textrm{DecBody}(\cdot)$ used for Eq.~\eqref{eq:decbody} consists of two attention modules:
\begin{align}
\nonumber
    Y_j[t]' &= Y_j[t] + \mathrm{MHA}^\mathrm{self}_j(Y_j[t], Y_j[1:t], Y_j[1:t]), \\
\nonumber
    Y_j'' &= Y_j + \mathrm{MHA}^\mathrm{src}_j(Y'_j, X_e, X_e), \\
    Y_{j+1} &= Y_j'' + \mathrm{FF}_j(Y_j''),
\end{align}
where $j = 0, \dots, d-1$ is the index of the decoder layers. 
We refer to the attention matrix between the decoder input and the encoder output in $\mathrm{MHA}^\mathrm{src}_j(Y'_j, X_e, X_e)$ as ``encoder-decoder attention' as same as the one in RNN in Sec~\ref{sec:rnn_dec}.
Because the unidirectional decoder is useful for sequence generation, its attention matrices at the $t$-th target frame are masked so that they do not connect with future frames later than $t$.
This masking of the sequence can be done in parallel using an elementwise product with a triangular binary matrix. Because it requires no sequential operation, it provides a faster implementation than RNN.


\subsection{Positional encoding}

To represent the time location in the non-recurrent model, Transformer adopts sinusoidal positional encoding:
\begin{align}
    \mathrm{PE}[t] = 
    \begin{cases}
    \sin{\frac{t}{10000^{t/d^\mathrm{att}}}} & \text{if $t$ is even}, \\
    \cos{\frac{t}{10000^{t/d^\mathrm{att}}}} & \text{if $t$ is odd}.
    \end{cases}
\end{align}
The input sequences $X_0, Y_0$ are concatenated with $(\textrm{PE}[1], \textrm{PE}[2], \dots)$  before $\textrm{EncBody}(\cdot)$ and  $\textrm{DecBody}(\cdot)$ modules. 

\section{ASR extensions}

In our ASR framework, the S2S predicts a target sequence $Y$ of characters or SentencePiece~\cite{kudo-richardson-2018-sentencepiece} from an input sequence $X^\mathrm{fbank}$ of log-mel filterbank speech features. 

\subsection{ASR encoder architecture}

The source $X$ in ASR is represented as a sequence of 83-dim log-mel filterbank frames with pitch features~\cite{kaldi-pitch}.
First, $\textrm{EncPre}(\cdot)$ transforms the source sequence $X$ into a subsampled sequence $X_0 \in \mathbb{R}^{n^\textrm{sub} \times d^\textrm{att}}$ by using two-layer CNN with 256 channels, stride size 2 and kernel size 3 in~\cite{speech-transformer}, or VGG-like max pooling in~\cite{HoriWZC17}, where $n^\mathrm{sub}$ is the length of the output sequence of the CNN. This CNN corresponds to $\textrm{EncPre}(\cdot)$ in Eq. (\ref{eq:encpre}). Then,  $\textrm{EncBody}(\cdot)$ transforms $X_0$ into a sequence of encoded features $X_e \in \mathbb{R}^{n^\mathrm{sub} \times d^\mathrm{att}}$ for the CTC and decoder networks.

\subsection{ASR decoder architecture}

The decoder network receives the encoded sequence $X_e$ and the prefix of a target sequence $Y[1:t-1]$ of  token IDs: characters or SentencePiece~\cite{kudo-richardson-2018-sentencepiece}. 
First, $\textrm{DecPre}(\cdot)$ in Eq.~\eqref{eq:decpre} embeds the tokens into learnable vectors. Next, $\textrm{DecBody}(\cdot)$ and single-linear layer $\textrm{DecPost}(\cdot)$ predicts the posterior distribution of the next token prediction $Y_\textrm{post}[t]$ given $X_e$ and $Y[1:t-1]$.


\subsection{ASR training and decoding}
\label{sec:asr_train_decode}
During ASR training, both the decoder and the CTC module predict the frame-wise posterior distribution of $Y$ given corresponding source $X$: $ p_\mathrm{s2s}(Y | X)$ and $p_\mathrm{ctc}(Y | X)$,  respectively. We simply use the weighted sum of those negative log likelihood values:
\begin{align}
L^\mathrm{ASR} = - \alpha \log p_\mathrm{s2s}(Y | X)- (1 - \alpha) \log p_\mathrm{ctc}(Y | X),
\end{align}
where $\alpha$ is a hyperparameter.

In the decoding stage, the decoder predicts the next token given the speech feature $X$ and the previous predicted tokens using beam search, which combines the scores of S2S, CTC and the RNN language model (LM)~\cite{Mikolov2010} as follows:
\begin{align}
\nonumber
\hat{Y} =  \mathop \mathrm{argmax} \limits_{Y \in \mathcal{Y}^*} \,\{\lambda \log p_{\mathrm{s2s}} (Y|X_e) &+ (1 - \lambda)\log p_{\mathrm{ctc}} (Y|X_e) \\ 
&+ \gamma \log p_{\mathrm{lm}} (Y)\},
\end{align}
where $\mathcal{Y}^*$ is a set of hypotheses of the target sequence, and $\gamma, \lambda$ are hyperparameters.

\section{ST extensions}

In ST, S2S receives the same source speech feature and target token sequences in ASR but the source and target languages are different.
Its modules are also defined in the same ways as in ASR. However, ST cannot cooperate with the CTC module introduced in Section~\ref{sec:asr_train_decode} because the translation task does not guarantee the monotonic alignment of the source and target sequences unlike ASR~\cite{Weiss2017}.

\section{TTS extensions}
In the TTS framework, the S2S generates a sequence of log-mel filterbank features and predicts the probabilities of the end of sequence (EOS) given an input character sequence~\cite{DBLP:conf/icassp/ShenPWSJYCZWRSA18}.

\subsection{TTS encoder architecture}
The input of the encoder in TTS is a sequence of IDs corresponding to the input characters and the EOS symbol.
First, the character ID sequence is converted into a sequence of character vectors with an embedding layer, and then the positional encoding scaled by a learnable scalar parameter is added to the vectors~\cite{li2019close}.
This process is a TTS implementation of $\textrm{EncPre}(\cdot)$ in Eq.~\eqref{eq:encpre}.
Finally, the encoder $\textrm{EncBody}(\cdot)$ in Eq.~\eqref{eq:encbody} transforms this input sequence into a sequence of encoded features for the decoder network.

\subsection{TTS decoder architecture}
The inputs of the decoder in TTS are a sequence of encoder features and a sequence of log-mel filterbank features. 
In training, ground-truth log-mel filterbank features are used with an teacher-forcing manner while in inference, predicted ones are used with an autoregressive manner.

First, the target sequence of 80-dim log-mel filterbank features is converted into a sequence of hidden features by Prenet~\cite{DBLP:conf/icassp/ShenPWSJYCZWRSA18} as a TTS implementation of $\textrm{DecPre}(\cdot)$ in Eq. (\ref{eq:decpre}). 
This network consists of two linear layers with 256 units, a ReLU activation function, and dropout followed by a projection linear layer with $d^\textrm{att}$ units.
Since it is expected that the hidden representations converted by Prenet are located in the similar feature space to that of encoder features, Prenet helps to learn a diagonal encoder-decoder attention~\cite{li2019close}.
Then the decoder $\textrm{DecBody}(\cdot)$ in Eq.~\eqref{eq:decbody}, whose architecture is the same as the encoder, transforms the sequence of encoder features and that of hidden features into a sequence of decoder features.
Two linear layers are applied for each frame of $Y_d$ to calculate the target feature and the probability of the EOS, respectively.
Finally, Postnet~\cite{DBLP:conf/icassp/ShenPWSJYCZWRSA18} is applied to the sequence of predicted target features to predict its components in detail. 
Postnet is a five-layer CNN, each layer of which is a 1d convolution with 256 channels and a kernel size of 5 followed by batch normalization, a tanh activation function, and dropout.
These modules are a TTS implementation of $\textrm{DecPost}(\cdot)$ in Eq.~\eqref{eq:post}.

\subsection{TTS training and decoding}
In TTS training, the whole network is optimized to minimize two loss functions in TTS; 1) L1 loss for the target features and 2) binary cross entropy (BCE) loss for the probability of the EOS.
To address the issue of class imbalance in the calculation of the BCE, a constant weight (e.g. 5) is used for a positive sample~\cite{li2019close}.

Additionally, we apply a guided attention loss~\cite{tachibana2018efficiently} to accelerate the learning of diagonal attention to only the two heads of two layers from the target side. This is because it is known that the encoder-decoder attention matrices are diagonal in only certain heads of a few layers from the target side~\cite{li2019close}.
We do not introduce any hyperparameters to balance the three loss values. We simply add them all together.

In inference, the network predicts the target feature of the next frame in an autoregressive manner. 
And if the probability of the EOS becomes higher than a certain threshold (e.g. 0.5), the network will stop the prediction.

\section{ASR Experiments}
\vspace{-2mm}
\subsection{Dataset}

In Table~\ref{tab:dataset}, we summarize the 15 datasets we used in our ASR experiment.
Our experiment covered various topics in ASR including recording (clean, noisy, far-field, etc), language (English, Japanese, Mandarin Chinese, Spanish, Italian) and size (10 - 960 hours).
Except for JSUT~\cite{jsut} and Fisher-CALLHOME Spanish, our data preparation scripts are based on Kaldi's ``s5x'' recipe~\cite{kaldi}.
Technically, we tuned all the configurations (e.g., feature extraction, SentencePiece~\cite{kudo-richardson-2018-sentencepiece}, language modeling, decoding, data augmentation~\cite{park2019specaugment,ko2015audio}) except for the training stage to their optimum in the existing RNN-based system.
We used data augmentation for several corpora. For example, we applied speed perturbation~\cite{ko2015audio} at ratio 0.9, 1.0 and 1.1 to CSJ, CHiME4, Fisher-CALLHOME Spanish, HKUST, and TED-LIUM2/3, and we also applied SpecAugment~\cite{park2019specaugment} to Aurora4, LibriSpeech, TED-LIUM2/3 and WSJ.\footnote{We chose datasets to apply these data augmentation methods by preliminary experiments with our RNN-based system.}

\begin{table*}[tb]
    \renewcommand{\arraystretch}{0.92}
    \caption{ASR dataset description. Names listed in ``test sets'' correspond to ASR results in Table~\ref{tab:asr}. We enlarged corpora marked with (*) by the external WSJ train\_si284 dataset (81 hours).}
    \centering
    \small
    \scalebox{0.87}{
    \begin{tabular}{c|ccccc}
    \toprule
    dataset & language & hours & speech & test sets \\
    \midrule
    AISHELL~\cite{aishell}          & zh & 170 & read & dev / test \\
    AURORA4~\cite{pearce2002aurora} (*) & en & 15 & noisy read & (dev\_0330) A / B / C / D \\    
    CSJ~\cite{CSJ-L00-1200}         & ja & 581 & spontaneous & eval1 / eval2 / eval3 \\
    CHiME4~\cite{chime3} (*)       & en & 108 & noisy far-field multi-ch read & dt05\_simu / dt05\_real / et05\_simu / et05\_real \\  
    CHiME5~\cite{chime5}            & en &  40 & noisy far-field multi-ch conversational & dev\_worn / kinect \\
    Fisher-CALLHOME Spanish        & es &  170 & telephone conversational & dev / dev2 / test / devtest / evltest \\
    HKUST~\cite{hkust}              & zh & 200 & telephone conversational & dev \\
    JSUT~\cite{jsut}                & ja &  10 & read & (our split) \\
    LibriSpeech~\cite{LibriSpeech}  & en & 960 & clean/noisy read & dev\_clean / dev\_other / test\_clean / test\_other \\ 
    REVERB~\cite{reverb} (*)        & en & 124 & far-field multi-ch read & et\_near / et\_far \\ 
    SWITCHBOARD~\cite{swbd}                & en & 260 & telephone conversational & (eval2000) callhm / swbd \\
    TED-LIUM2~\cite{TED-LIUM/ROUSSEAU12.698} & en & 118 & spontaneous & dev / test \\
    TED-LIUM3~\cite{tedlium3}        & en & 452 & spontaneous & dev / test \\
    VoxForge~\cite{voxforge}    & it &  16 & read & (our split) \\
    WSJ~\cite{wsjPaul:1992:DWS:1075527.1075614}        & en &  81 & read & dev93 / eval92 \\
     \bottomrule
    \end{tabular}
    }
    \label{tab:dataset}
    \vspace{-5mm}
    \renewcommand{\arraystretch}{1.0}
\end{table*}

\subsection{Settings}
\label{sec:asr_setting}
We adopted the same architecture for Transformer in~\cite{Karita2019} ($e = 12, d = 6, d^\textrm{ff} = 2048, d^\textrm{head} = 4, d^\textrm{att} = 256$) for every corpus except for the largest, LibriSpeech ($d^\textrm{head} = 8, d^\textrm{att} = 512$).
For RNN, we followed our existing best architecture configured on each corpus as in previous studies~\cite{hori2018end,Zeyer2018}.

Transformer requires a different optimizer configuration from RNN because Transformer's training iteration is eight times faster and its update is more fine-grained than RNN. 
For RNN, we followed existing best systems for each corpus using Adadelta~\cite{adadelta} with early stopping. To train Transformer, we basically followed the previous literature~\cite{speech-transformer} (e.g., dropout, learning rate, warmup steps). 
We did not use development sets for early stopping in Transformer. We simply ran 20 -- 200 epochs (mostly 100 epochs) and averaged the model parameters stored at the last 10 epochs as the final model.  

We conducted our training on a single GPU for larger corpora such as LibriSpeech, CSJ and TED-LIUM3. We also confirmed that the emulation of multiple GPUs using accumulating gradients over multiple forward/backward steps~\cite{ott-etal-2018-scaling} could result in similar performance with those corpora.
In the decoding stage, Transformer and RNN share the same configuration for each corpus, for example, beam size (e.g., 20 -- 40), CTC weight $\lambda$ (e.g., 0.3), and LM weight $\gamma$ (e.g., 0.3 -- 1.0) introduced in Section \ref{sec:asr_train_decode}.

\subsection{Results}

Table~\ref{tab:asr} summarizes the ASR results in terms of character/word error rate (CER/WER) on each corpora.
It shows that Transformer outperforms RNN on 13/15 corpora in our experiment.
Although our system has no pronunciation dictionary, part-of-speech tag nor alignment-based data cleaning unlike Kaldi,
our Transformer provides comparable CER/WERs to the HMM-based system, Kaldi on 7/12 corpora.
We conclude that Transformer has ability to outperform the RNN-based end-to-end system and the DNN/HMM-based system even in low resource (JSUT), large resource (LibriSpeech, CSJ), noisy (AURORA4) and far-field (REVERB) tasks.
Table~\ref{tab:libri_asr} also summarizes the LibriSpeech ASR benchmark with ours and other reports because it is one of the most competitive task. Our transformer results are comparable to the best performance in~\cite{irie2019language,luscher2019rwth,park2019specaugment}.

\begin{figure}
    \centering
    \includegraphics[width=\columnwidth]{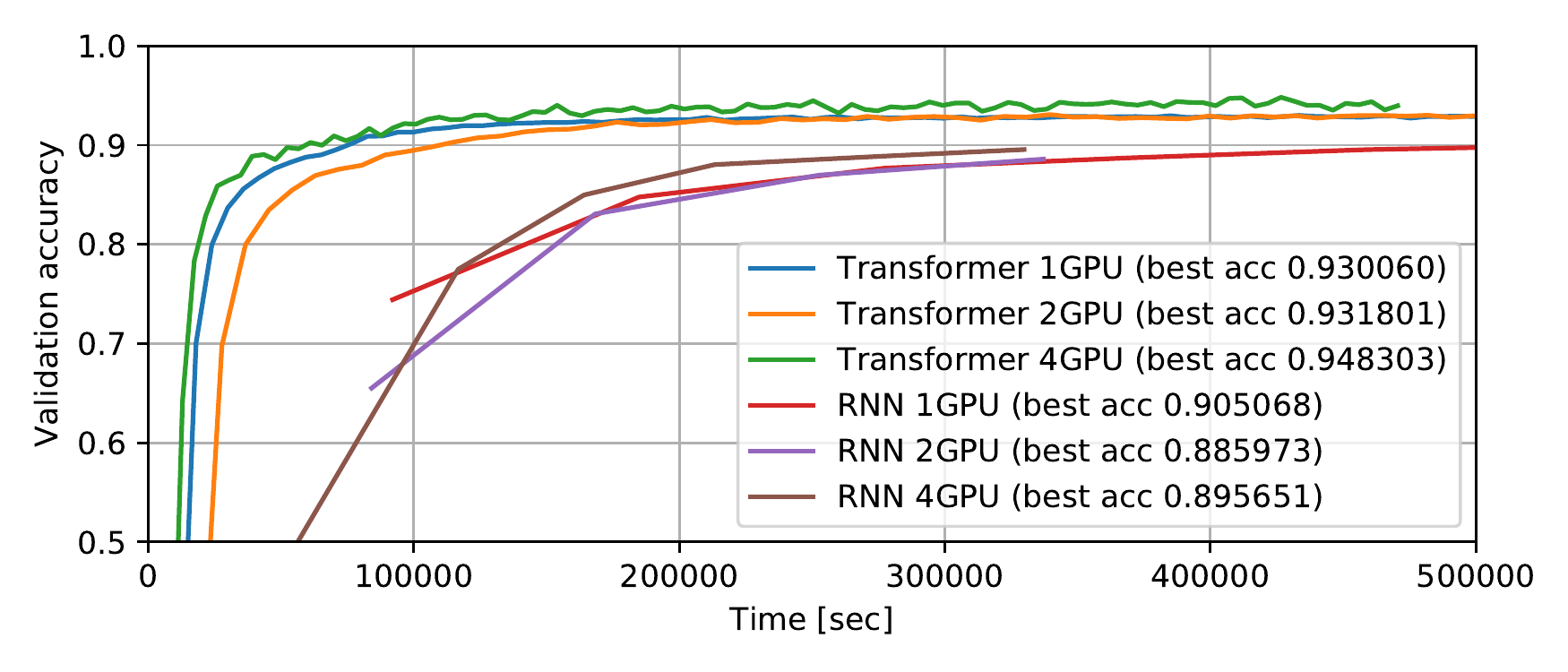}
    \vspace{-10mm}
\caption{ASR training curve with LibriSpeech dataset. Minibatches had the maximum number of utterances for each models on GPUs.}
    \vspace{-5mm}
    \label{fig:asr_training}
\end{figure}

Fig.~\ref{fig:asr_training} shows an ASR training curve obtained with multiple GPUs on LibriSpeech.
We observed that Transformer trained with a larger minibatch became more accurate while RNN did not.
On the other hand, when we use a smaller minibatch for Transformer, it typically became under-fitted after the warmup steps.
In this task, Transformer achieved the best accuracy provided by RNN about eight times faster than RNN with a single GPU.

\begin{table*}[tb]
  \renewcommand{\arraystretch}{0.92}
  \caption{ASR results of char/word error rates. Results marked with (*) were evaluated in our environment because the official results were not provided. Kaldi official results were retrieved from the version ``c7876a33''.}
  \label{tab:asr}
  \centering
  \small
  \scalebox{0.87}{
  \begin{tabular}{ ccc | r | r r r r }
    \toprule
    \multicolumn{1}{c}{{dataset}} & 
    \multicolumn{1}{c}{{token}} & 
    \multicolumn{1}{c}{{error}} & 
    \multicolumn{1}{c}{{Kaldi (s5)}} & 
    \multicolumn{1}{c}{{ESPnet RNN (ours)}} &
    \multicolumn{1}{c}{{ESPnet Transformer (ours)}} \\
    \midrule
    AISHELL     & char & CER & N/A / 7.4 & 6.8 / 8.0 & \textbf{6.0} / \textbf{6.7} \\
    AURORA4     & char & WER 
                & (*) 3.6 / 7.7 / 10.0 / 22.3
                & 3.5 / 6.4 / 5.1 / 12.3
                & \textbf{3.3} / \textbf{6.0} / \textbf{4.5} / \textbf{10.6} \\
    CSJ         & char & CER
                & (*) 7.5 / 6.3 / 6.9  
                & 6.6 / 4.8 / 5.0 
                & \textbf{5.7} / \textbf{4.1} / \textbf{4.5} \\
    CHiME4      
                & char & WER
                & \textbf{6.8} / \textbf{5.6} / \textbf{12.1} / \textbf{11.4}
                & 9.5 / 8.9 / 18.3 / 16.6
                & 9.6 / 8.2 / 15.7 / 14.5 \\
    CHiME5      & char & WER
                & \textbf{47.9} / \textbf{81.3}
                & 59.3 / 88.1
                & 60.2 / 87.1 \\
    Fisher-CALLHOME Spanish & char & WER
                & N/A
                & 27.9 / 27.8 / 25.4 / 47.2 / 47.9
                & \textbf{27.0} / \textbf{26.3} / \textbf{24.4} / \textbf{45.3} / \textbf{46.2}  \\
    HKUST       & char & CER & 23.7 & 27.4 & \textbf{23.5} \\
    JSUT        & char & CER & N/A  & 20.6 & \textbf{18.7} \\
    LibriSpeech & BPE  & WER 
                & 3.9 / 10.4 / 4.3 / 10.8
                & 3.1 / 9.9 / 3.3 / 10.8  
                & \textbf{2.2} / \textbf{5.6} / \textbf{2.6} / \textbf{5.7} \\
    REVERB      & char & WER
                & 18.2 / 19.9 
                & 24.1 / 27.2
                & \textbf{15.5} / \textbf{19.0} \\
    SWITCHBOARD & BPE & WER & \textbf{18.1} / \textbf{8.8}  & 28.5 / 15.6 & \textbf{18.1} / 9.0 \\
    TED-LIUM2    
                & BPE  & WER &      \textbf{9.0} / 9.0 & 11.2 / 11.0 & 9.3 / \textbf{8.1} \\
    TED-LIUM3    & BPE  & WER & \textbf{6.2} / \textbf{6.8} & 14.3 / 15.0 &  9.7 / 8.0  \\
    VoxForge    & char & CER & N/A & 12.9 / 12.6 & \textbf{9.4} / \textbf{9.1} \\
    WSJ         
                & char & WER & \textbf{4.3} / \textbf{2.3} & 7.0 / 4.7 & 6.8 / 4.4 \\
    \bottomrule
  \end{tabular}
  }
  \vspace{-2mm}
  \renewcommand{\arraystretch}{1.0}
\end{table*}
\begin{table}[tb]
  \caption{Comparison of the Librispeech ASR benchmark}
  \label{tab:libri_asr}
  \centering
  \scalebox{0.85}{
  \begin{tabular}{ c | r r r r }
  \toprule
        & dev\_clean    & dev\_other    & test\_clean   & test\_other \\
  \midrule
  RWTH (E2E) \cite{irie2019language}       & 2.9           & 8.8           & 3.1           & 9.8\\
  RWTH (HMM) \cite{luscher2019rwth}        & 2.3           & \textbf{5.2}  & 2.7           & \textbf{5.7}\\
  Google SpecAug. \cite{park2019specaugment} & N/A           & N/A           & \textbf{2.5}  & 5.8\\
  ESPnet Transformer (ours) & \textbf{2.2}  & 5.6           & 2.6           & \textbf{5.7}\\
  \bottomrule
  \end{tabular}
  }
  \vspace*{-5mm}
\end{table}

\subsection{Discussion}

We summarize the training tips we observed in our experiment:
{
\setlength{\leftmargini}{15pt}  
\begin{itemize}
	\setlength{\itemsep}{1pt}      
	\setlength{\parskip}{0pt}      
	\setlength{\itemindent}{0pt}   
	\setlength{\labelsep}{4pt}     
    \item When Transformer suffers from under-fitting, we recommend increasing the minibatch size because it also results in a faster training time and better accuracy simultaneously unlike any other hyperparameters.
    \item The accumulating gradient strategy~\cite{ott-etal-2018-scaling} can be adopted to emulate the large minibatch if multiple GPUs are unavailable.
    \item While dropout did not improve the RNN results, it is essential for Transformer to avoid over-fitting.
    \item We tried several data augmentation methods~\cite{ko2015audio,park2019specaugment}. They greatly improved both Transformer and RNN.
    \item The best decoding hyperparameters $\gamma, \lambda$ for RNN are generally the best for Transformer.
\end{itemize}
}
Transformer's weakness is decoding. It is much slower than Kaldi's system because the self-attention requires $O(n^2)$ in a naive implementation, 
where $n$ is the speech length. To directly compare the performance with DNN-HMM based ASR systems, we need to develop a faster decoding algorithm for Transformer.

\section{Multilingual ASR Experiments}
This section compares the ASR performance of RNN and Transformer in a multilingual setup given the success of Transformer for the monolingual ASR tasks in the previous section.
In accordance with~\cite{watanabe2017language}, we prepared 10 different languages, namely WSJ (English), CSJ (Japanese) \cite{CSJ-L00-1200}, HKUST \cite{hkust} (Mandarin Chinese), and VoxForge (German, Spanish, French, Italian, Dutch, Portuguese, Russian).
The model is based on a single multilingual model,  where the parameters
are shared across all the languages and whose output units include the graphemes of all 10 languages (totally 5,297 graphemes and special symbols).
We used a default setup for both RNN and Transformer introduced in Section \ref{sec:asr_setting} without RNNLM shallow fusion \cite{HoriWZC17}.
\begin{figure}[tb]
    \centering
    \vspace{-3mm}
    \includegraphics[width=0.9\columnwidth]{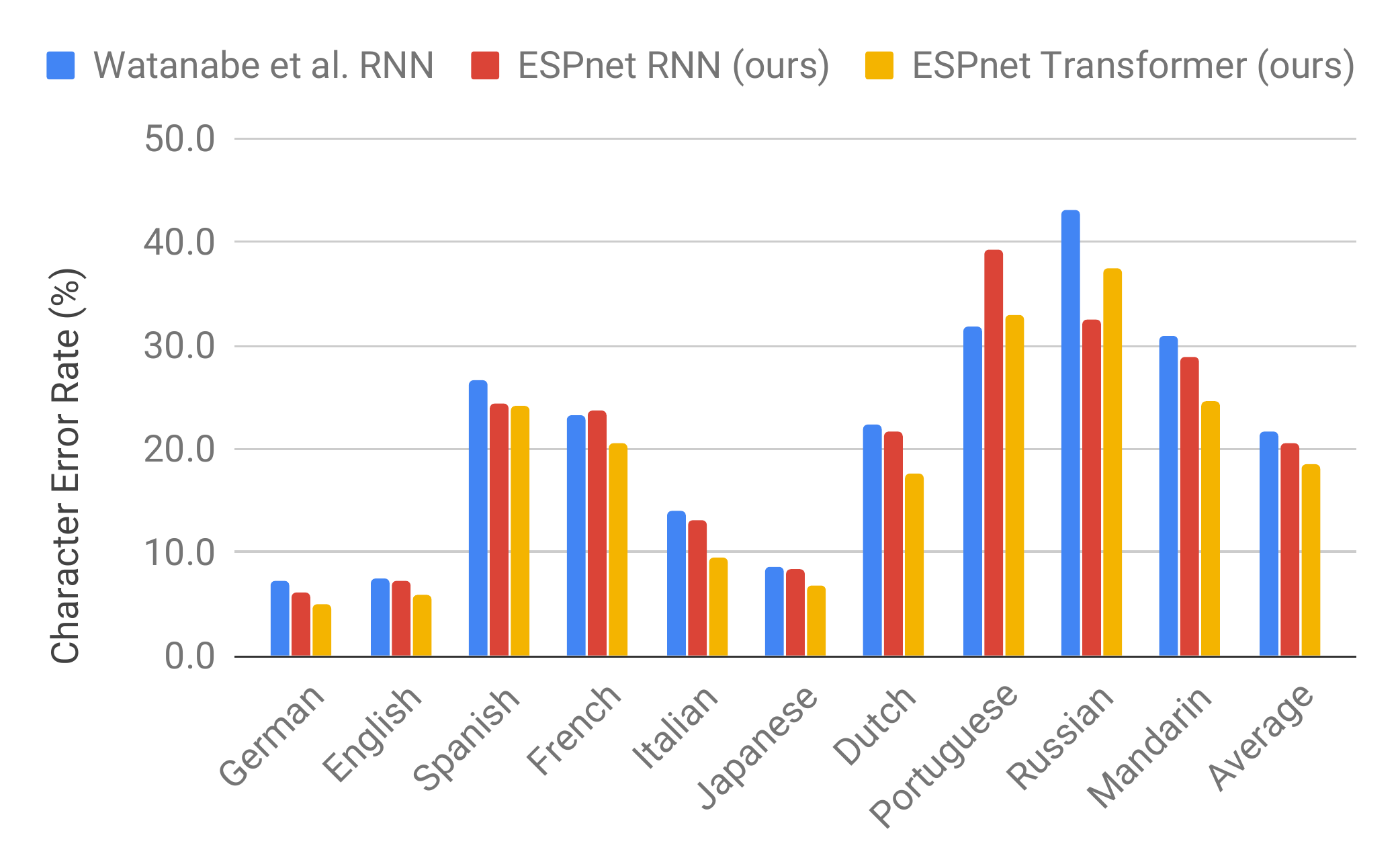}
    \vspace{-6mm}
    \caption{Comparison of multilingual end-to-end ASR with the RNN in Watanabe et al.~\cite{watanabe2017language}, ESPnet RNN, and ESPnet Transformer.}
    \vspace{-2mm}
    \label{fig:li10}
\end{figure}

Figure \ref{fig:li10} clearly shows that our Transformer significantly outperformed our RNN in 9 languages.
It realized a more than 10\% relative improvement in 8 languages and with the largest value of 28.0\% for relative improvement in VoxForge Italian.
When compared with the RNN result reported in \cite{watanabe2017language}, which used a deeper BLSTM (7 layer) and RNNLM, our Transformer still provided superior performance in 9 languages. 
From this result, we can conclude that Transformer also outperforms RNN in multilingual end-to-end ASR.

\section{Speech Translation Experiments}

Our baseline end-to-end ST RNN is based on~\cite{Weiss2017}, which is similar to the RNN structure used in our ASR system, but we did not use a convolutional LSTM layer in the original paper.
The configuration of our ST Transformer was the same as that of our ASR system.

We conducted our ST experiment on the Fisher-CALLHOME English--Spanish corpus~\cite{post2013improved}.
Our Transformer improved the BLEU score to 17.2 from our RNN baseline BLEU 16.5 on the CALLHOME ``evltest'' set. 
While training Transformer, we observed more serious under-fitting than with RNN.
The solution for this is to use the pretrained encoder from our ASR experiment since the ST dataset contains Fisher-CALLHOME Spanish corpus used in our ASR experiment.

\section{TTS Experiments}

\subsection{Settings}
Our baseline RNN-based TTS model is Tacotron 2~\cite{DBLP:conf/icassp/ShenPWSJYCZWRSA18}.
We followed its model and optimizer setting.
We reuse existing TTS recipes including those for data preparation and waveform generation that we configured to be the best for RNN.
We configured our Transformer-based configurations introduced in Section \ref{sec:transformer} as follows: $e=6, d=6, d^\textrm{att}=384, d^\textrm{ff}=1536, d^\textrm{head}=4$. 
\tomokiedit{The input for both systems was the sequence of characters.}

\subsection{Results}
\begin{figure}[t!]
    \centering
    \includegraphics[width=1\columnwidth]{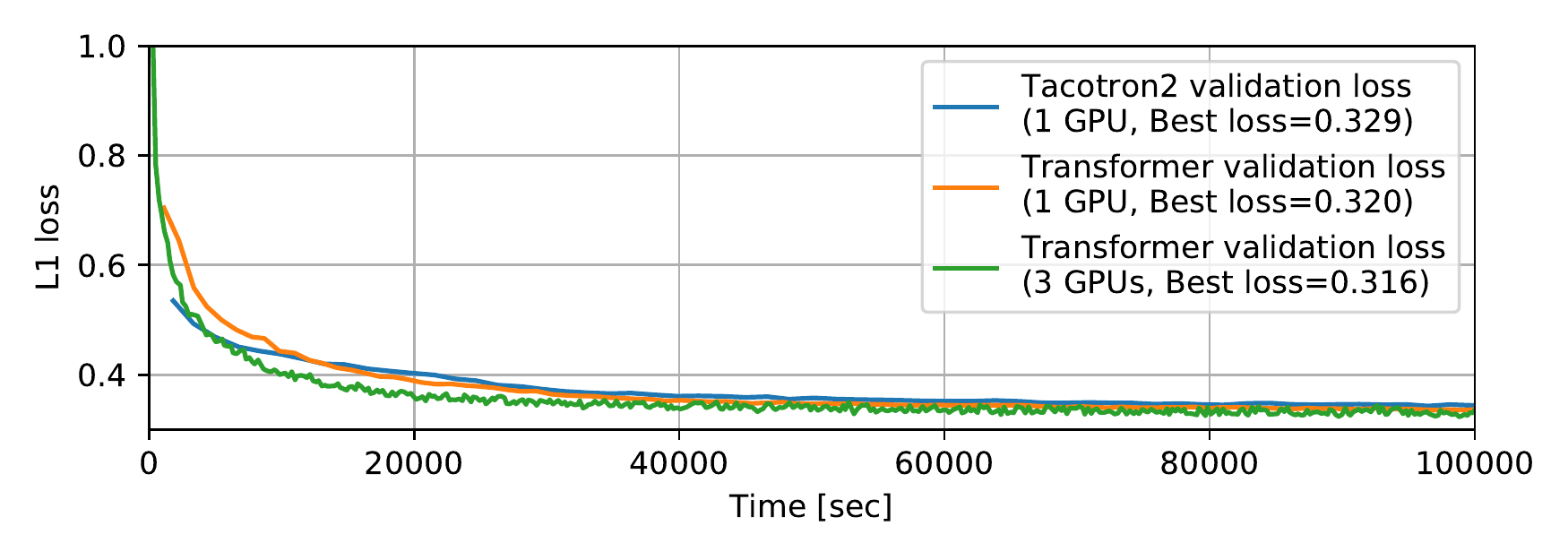}
    \vspace{-10mm}
    \caption{TTS training curve on M-AILABS.}
    \vspace{-5mm}
    \label{fig:training_curve1}
\end{figure}

\begin{figure}[t!]
    \centering
    \includegraphics[width=1\columnwidth]{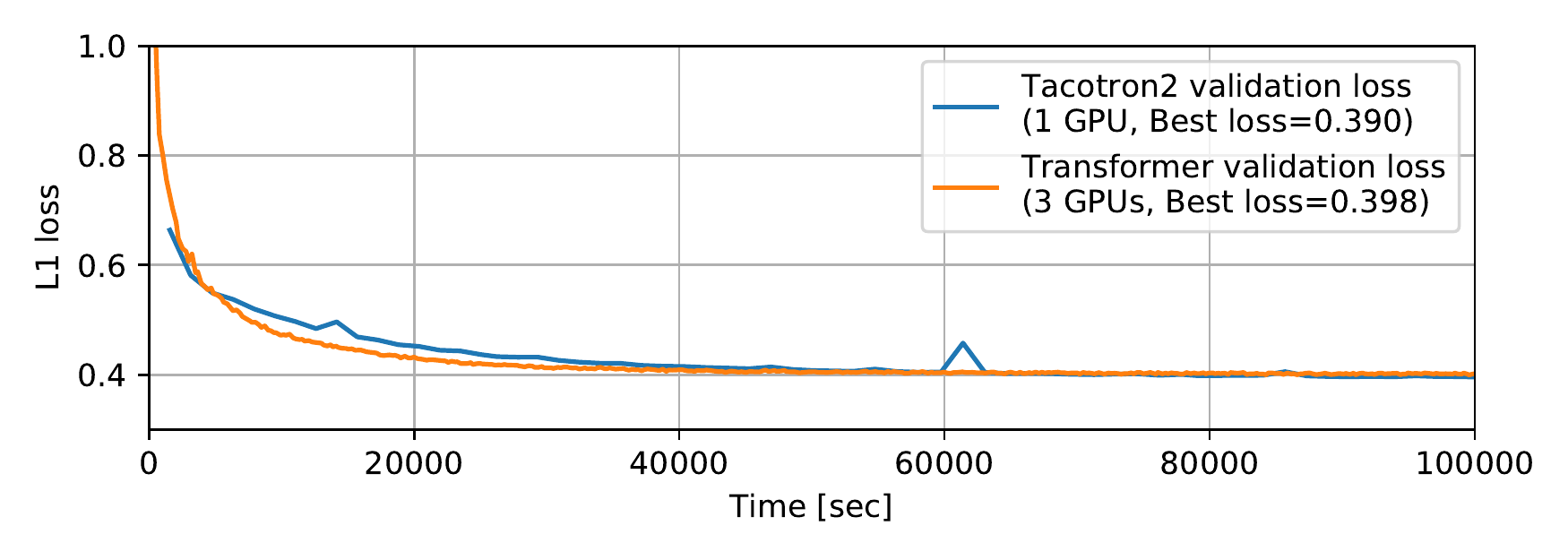}
    \vspace{-10mm}
    \caption{TTS training curve on LJSpeech.}
    \vspace{0mm}
    \label{fig:training_curve2}
\end{figure}

\begin{figure}[t!]
    \centering
    \includegraphics[width=1\columnwidth]{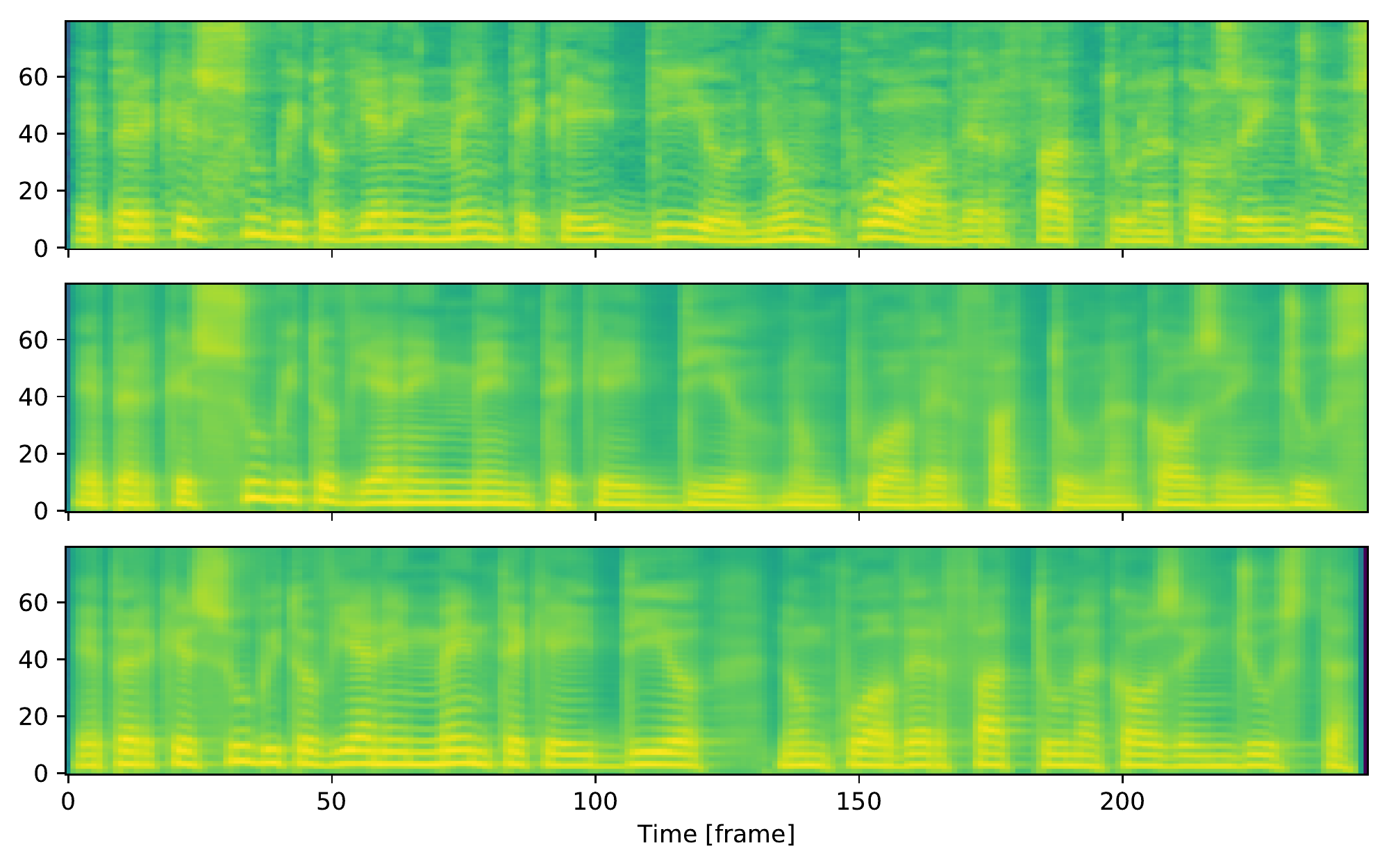}
    \vspace{-10mm}
    \caption{Samples of mel-spectrograms on M-AILABs. (top) ground-truth, (middle) Tacotron 2 sample, (bottom) Transformer sample. The input text is {\it ``E PERCHÈ SUBITO VIENE IN MENTE CHE IDDIO NON PUÒ AVER FATTO UNA COSA INGIUSTA''}.}
    \label{fig:tts_sample1}
    \vspace{-3mm}
\end{figure}

\begin{figure}[t!]
    \centering
    \includegraphics[width=1\columnwidth]{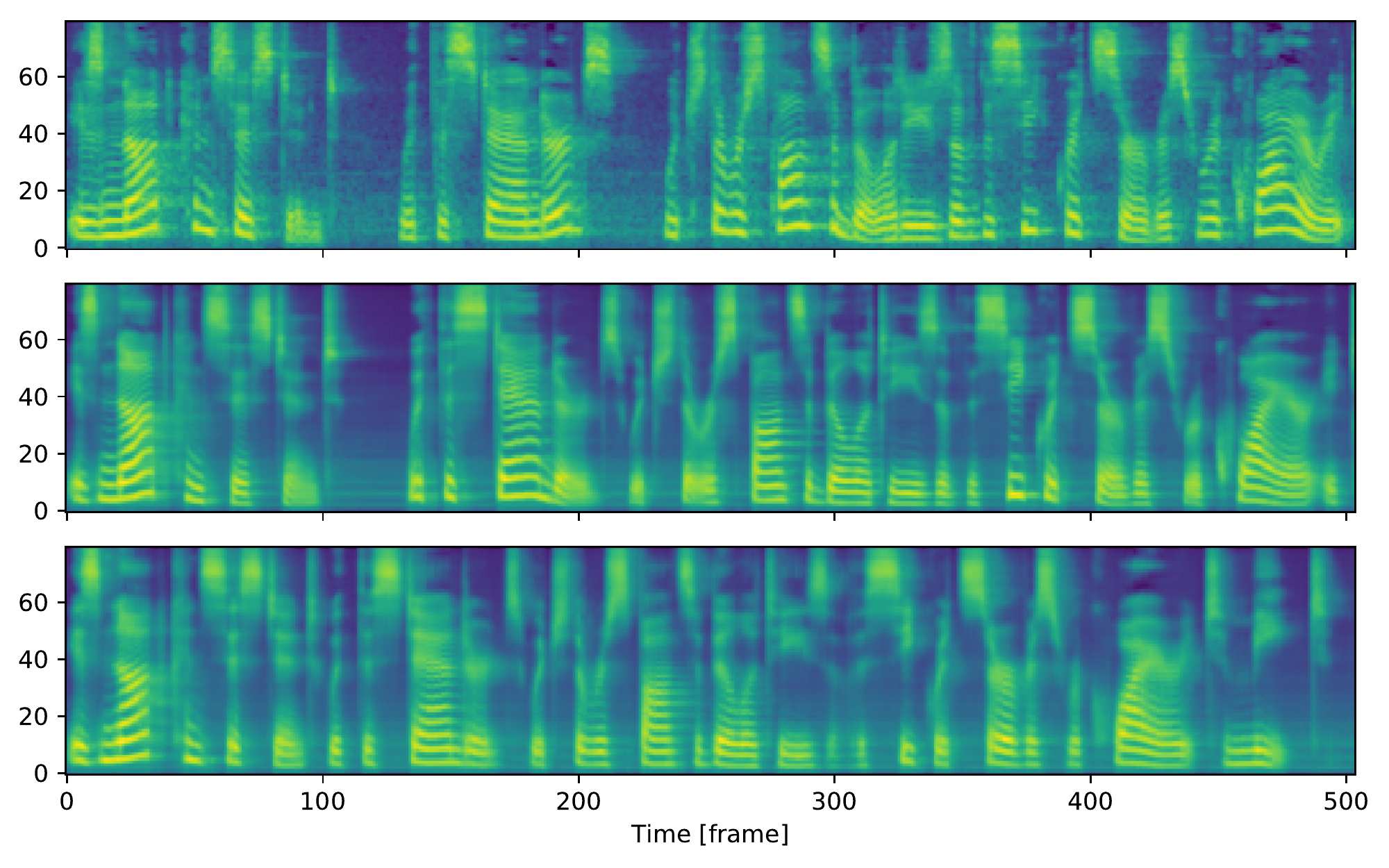}
    \vspace{-10mm}
    \caption{Samples of mel-spectrograms on LJSpeech. (top) ground-truth, (middle) Tacotron 2 sample, (bottom) Transformer sample. The input text is {\it ``IS NOT CONSISTENT WITH THE STANDARDS WHICH THE RESPONSIBILITIES OF THE SECRET SERVICE REQUIRE IT TO MEET.''.}}
    \label{fig:tts_sample2}
    \vspace{-5mm}
\end{figure}

We compared Transformer and RNN based TTS using two corpora: M-AILABS~\cite{mailabs} \tomokiedit{(Italian, 16 kHz, 31 hours)} and LJSpeech~\cite{ljspeech17} \tomokiedit{(English, 22 kHz, 24 hours)}.
\tomokiedit{A single Italian male speaker (Riccardo) was used in the case of M-AILABS.}
Figures ~\ref{fig:training_curve1} and \ref{fig:training_curve2} show training curves in the two corpora.
In these figures, Transformer and RNN provide similar L1 loss convergence.
As seen in ASR, we observed that a larger minibatch results in better validation L1 loss for Transformer and faster training, while it has a detrimental effect on the L1 loss for RNN.
We also provide generated speech mel-spectrograms in Fig.~\ref{fig:tts_sample1} and \ref{fig:tts_sample2}\footnote{Our audio samples generated by Tacotron 2, Transformer, and FastSpeech are available at \url{https://bit.ly/329gif5}}.
We conclude that Transformer-based TTS can achieve almost the same performance as RNN-based.


\subsection{Discussion}
Our lessons for training Transformer in TTS are as follows:
{
\setlength{\leftmargini}{15pt}  
\begin{itemize}
	\setlength{\itemsep}{1pt}      
	\setlength{\parskip}{0pt}      
	\setlength{\itemindent}{0pt}   
	\setlength{\labelsep}{4pt}     
    \item It is possible to accelerate TTS training by using a large minibatch as well as ASR if a lot of GPUs are available.
    \item The validation loss value\tomokiedit{, especially BCE loss,} could be over-fitted more easily with Transformer. We recommend monitoring attention maps rather than the loss when checking its convergence.
    \item Some heads of attention maps in Transformer are not always diagonal as found with Tacotron 2. We needed to \tomokiedit{select} where to apply the guided attention loss~\cite{tachibana2018efficiently}.
    \item Decoding filterbank features with Transformer is also slower than with RNN \tomokiedit{(6.5 ms vs 78.5 ms per frame, on CPU w/ single thread)}. We also tried FastSpeech~\cite{fastspeech}, which realizes non-autoregressive Transformer-based TTS. It greatly improves the decoding speed \tomokiedit{(0.6 ms per frame, on CPU w/ single thread)} and generates comparable quality of speech with the autoregressive Transformer.
    \item \tomokiedit{A reduction factor introduced in~\cite{Wang2017} was also effective for Transformer. It can greatly reduce training and inference time but slightly degrades the quality.}
\end{itemize}
}
As future work, we need further investigation of the trade off between training speed and quality, and the introduction of ASR techniques (e.g., data augmentation, speech enhancement) for TTS.


\section{Summary}

We presented a comparative study of Transformer and RNN in speech applications with various corpora, namely ASR (15 monolingual + one multilingual), ST (one corpus), and TTS (two corpora).
In our experiments on these tasks, we obtained the promising results including huge improvements in many ASR tasks and explained how we improved our models.
We believe that the reproducible recipes, pretrained models and training tips described in this paper will accelerate Transformer research directions on speech applications.


\section{References}
\printbibliography

@inproceedings{watanabe2017language,
  title={Language independent end-to-end architecture for joint language identification and speech recognition},
  author={Watanabe, Shinji and Hori, Takaaki and Hershey, John R},
  booktitle={IEEE Automatic Speech Recognition and Understanding Workshop (ASRU)},
  pages={265--271},
  year={2017},
  organization={IEEE}
}

@inproceedings{ko2015audio,
  title={Audio augmentation for speech recognition},
  author={Ko, Tom and Peddinti, Vijayaditya and Povey, Daniel and Khudanpur, Sanjeev},
  booktitle={Sixteenth Annual Conference of the International Speech Communication Association},
  year={2015}
}

@incollection{s2s_NIPS2014_5346,
title = {Sequence to Sequence Learning with Neural Networks},
author = {Sutskever, Ilya and Vinyals, Oriol and Le, Quoc V},
booktitle = {Advances in Neural Information Processing Systems 27},
editor = {Z. Ghahramani and M. Welling and C. Cortes and N. D. Lawrence and K. Q. Weinberger},
pages = {3104--3112},
year = {2014},
publisher = {Curran Associates, Inc.},
url = {http://papers.nips.cc/paper/5346-sequence-to-sequence-learning-with-neural-networks.pdf}
}

@inproceedings{park2019specaugment,
  author = {D. S. Park and W. Chan and Y. Zhang and C. Chiu and B. Zoph and E. D. Cubuk and Q. V. Le},
  title = {{SpecAugment}: A Simple Data Augmentation Method for Automatic Speech Recognition},
  booktitle = {arXiv},
  year = {2019}
}

@inproceedings{Karita2019,
  author={Shigeki Karita and Nelson Enrique Yalta Soplin and Shinji Watanabe and Marc Delcroix and Atsunori Ogawa and Tomohiro Nakatani},
  title={{Improving Transformer-Based End-to-End Speech Recognition with Connectionist Temporal Classification and Language Model Integration}},
  year=2019,
  booktitle={Proc. Interspeech 2019},
  pages={1408--1412},
  doi={10.21437/Interspeech.2019-1938},
  url={http://dx.doi.org/10.21437/Interspeech.2019-1938}
}

@inproceedings{kudo-richardson-2018-sentencepiece,
    title = "{S}entence{P}iece: A simple and language independent subword tokenizer and detokenizer for Neural Text Processing",
    author = "Kudo, Taku  and
      Richardson, John",
    booktitle = "Proceedings of the 2018 Conference on Empirical Methods in Natural Language Processing: System Demonstrations",
    month = nov,
    year = "2018",
    url = "https://www.aclweb.org/anthology/D18-2012",
    pages = "66--71",
}

@InProceedings{tedlium3,
author={Hernandez, Francois and Nguyen, Vincent and Ghannay, Sahar and Tomashenko, Natalia and Esteve, Yannick and Jokisch, Oliver and Potapova, Rodmonga},
title="{TED-LIUM} 3: Twice as Much Data and Corpus Repartition for Experiments on Speaker Adaptation",
booktitle="Speech and Computer",
year="2018",
publisher="Springer International Publishing",
address="Cham",
pages="198--208",
isbn="978-3-319-99579-3"
}

@misc{voxforge,
  title={{VoxForge}},
  howpublished = {\url{http://www.voxforge.org}},
}

@misc{ljspeech17,
  author       = {Keith Ito},
  title        = {The {LJ Speech} Dataset},
  howpublished = {\url{https://keithito.com/LJ-Speech-Dataset/}},
  year         = 2017
}

@misc{mailabs,
author = {Imdat Solak},
title = {The {M-AILABS} Speech Dataset},
howpublished = {\url{https://www.caito.de/2019/01/the-m-ailabs-speech-dataset/}},
year = 2019
}

@inproceedings{swbd,
  author = {Godfrey, J. and Holliman, E. and McDaniel, J.},
  booktitle = {IEEE International Conference on Speech, and Signal Processing, ICASSP-92},
  pages = {517-520},
  title = {{SWITCHBOARD}: telephone speech corpus for research and development},
  volume = 1,
  year = 1992
}

@inproceedings{wsjPaul:1992:DWS:1075527.1075614,
 author = {Paul, Douglas B. and Baker, Janet M.},
 title = {The Design for the {Wall Street Journal}-based {CSR} Corpus},
 booktitle = {Proceedings of the Workshop on Speech and Natural Language},
 series = {HLT '91},
 year = {1992},
 isbn = {1-55860-272-0},
 location = {Harriman, New York},
 pages = {357--362},
 numpages = {6},
 url = {https://doi.org/10.3115/1075527.1075614},
 doi = {10.3115/1075527.1075614},
 acmid = {1075614},
 publisher = {Association for Computational Linguistics},
 address = {Stroudsburg, PA, USA},
}

@InProceedings{reverb,
author="Kinoshita, Keisuke
and Delcroix, Marc
and Gannot, Sharon
and Habets, Emanuel A. P.
and Haeb-Umbach, Reinhold
and Kellermann, Walter
and Leutnant, Volker
and Maas, Roland
and Nakatani, Tomohiro
and Raj, Bhiksha
and Sehr, Armin
and Yoshioka, Takuya",
title="The {REVERB} Challenge: A Benchmark Task for Reverberation-Robust ASR Techniques",
bookTitle="New Era for Robust Speech Recognition: Exploiting Deep Learning",
year="2017",
publisher="Springer International Publishing",
address="Cham",
pages="345--354",
isbn="978-3-319-64680-0",
doi="10.1007/978-3-319-64680-0_15",
url="https://doi.org/10.1007/978-3-319-64680-0_15"
}

@article{jsut,
  author    = {Ryosuke Sonobe and
               Shinnosuke Takamichi and
               Hiroshi Saruwatari},
  title     = {{JSUT} corpus: free large-scale Japanese speech corpus for end-to-end
               speech synthesis},
  journal   = {CoRR},
  volume    = {abs/1711.00354},
  year      = {2017},
  url       = {http://arxiv.org/abs/1711.00354},
  archivePrefix = {arXiv},
  eprint    = {1711.00354},
  bibsource = {dblp computer science bibliography, https://dblp.org}
}

@inproceedings{chime5,
  author={Jon Barker and Shinji Watanabe and Emmanuel Vincent and Jan Trmal},
  title={The Fifth {'CHiME'} Speech Separation and Recognition Challenge: Dataset, Task and Baselines},
  year=2018,
  booktitle={Proc. Interspeech},
  pages={1561--1565},
  doi={10.21437/Interspeech.2018-1768},
  url={http://dx.doi.org/10.21437/Interspeech.2018-1768}
}

@article{chime3,
  title={The third {CHiME} speech separation and recognition challenge: Analysis and outcomes},
  author={Barker, Jon and Marxer, Ricard and Vincent, Emmanuel and Watanabe, Shinji},
  journal={Computer Speech \& Language},
  volume={46},
  pages={605--626},
  year={2017},
  publisher={Elsevier}
}

@InProceedings{hkust,
author="Liu, Yi
and Fung, Pascale
and Yang, Yongsheng
and Cieri, Christopher
and Huang, Shudong
and Graff, David",
title="{HKUST/MTS}: A Very Large Scale Mandarin Telephone Speech Corpus ",
booktitle="Chinese Spoken Language Processing",
year="2006",
publisher="Springer Berlin Heidelberg",
address="Berlin, Heidelberg",
pages="724--735",
isbn="978-3-540-49666-3"
}

@article{pearce2002aurora,
  title={Aurora working group: DSR front end LVCSR evaluation AU/384/02},
  author={Pearce, David and Picone, J},
  journal={Inst. for Signal \& Inform. Process., Mississippi State Univ., Tech. Rep},
  year={2002}
}

@INPROCEEDINGS{aishell, 
author={H. {Bu} and J. {Du} and X. {Na} and B. {Wu} and H. {Zheng}}, 
booktitle={2017 20th Conference of the Oriental Chapter of the International Coordinating Committee on Speech Databases and Speech I/O Systems and Assessment (O-COCOSDA)}, 
title={{AISHELL-1}: An open-source {Mandarin} speech corpus and a speech recognition baseline}, 
year={2017}, 
volume={}, 
number={}, 
pages={1-5}, 
doi={10.1109/ICSDA.2017.8384449}, 
ISSN={2472-7695}, }

@article{DBLP:journals/pbml/PopelB18,
  author    = {Martin Popel and
               Ondrej Bojar},
  title     = {Training Tips for the Transformer Model},
  journal   = {Prague Bull. Math. Linguistics},
  volume    = {110},
  pages     = {43--70},
  year      = {2018},
  url       = {http://ufal.mff.cuni.cz/pbml/110/art-popel-bojar.pdf},
  bibsource = {dblp computer science bibliography, https://dblp.org}
}

@inproceedings{ctc-DBLP:conf/icml/GravesFGS06,
  author    = {Alex Graves and
               Santiago Fern{\'{a}}ndez and
               Faustino J. Gomez and
               J{\"{u}}rgen Schmidhuber},
  title     = {Connectionist temporal classification: labelling unsegmented sequence
               data with recurrent neural networks},
  booktitle = {{ICML}},
  series    = {{ACM} International Conference Proceeding Series},
  volume    = {148},
  pages     = {369--376},
  publisher = {{ACM}},
  year      = {2006}
}

@InProceedings{luong-dot-att-D15-1166,
  author = 	"Luong, Thang
		and Pham, Hieu
		and Manning, Christopher D.",
  title = 	"Effective Approaches to Attention-based Neural Machine Translation",
  booktitle = 	"Proceedings of the 2015 Conference on Empirical Methods in Natural      Language Processing    ",
  year = 	"2015",
  publisher = 	"Association for Computational Linguistics",
  pages = 	"1412--1421",
  doi = 	"10.18653/v1/D15-1166",
  url = 	"http://aclweb.org/anthology/D15-1166"
}

@article{adadelta,
  author = {Zeiler, Matthew D.},
  ee = {http://arxiv.org/abs/1212.5701},
  journal = {CoRR},
  title = {ADADELTA: An Adaptive Learning Rate Method},
  url = {http://dblp.uni-trier.de/db/journals/corr/corr1212.html#abs-1212-5701},
  volume = {abs/1212.5701},
  year = 2012
}

@InProceedings{CSJ-L00-1200,
	author = 	"Maekawa, Kikuo
	and Koiso, Hanae
	and Furui, Sadaoki
	and Isahara, Hitoshi",
	title = 	"Spontaneous Speech Corpus of {Japanese}",
	booktitle = 	"Proceedings of the Second International Conference on Language Resources and Evaluation (LREC'00)",
	year = 	"2000",
	publisher = 	"European Language Resources Association (ELRA)",
	url = 	"http://www.lrec-conf.org/proceedings/lrec2000/pdf/262.pdf"
}

@InProceedings{TED-LIUM/ROUSSEAU12.698,
	author    = {Anthony Rousseau and
	Paul Deleglise and
	Yannick Esteve},
	title = {{TED-LIUM}: an automatic speech recognition dedicated corpus},
	booktitle = {Proceedings of the Eighth International Conference on Language Resources and Evaluation (LREC'12)},
	year = {2012},
	address = {Istanbul, Turkey},
	publisher = {European Language Resources Association (ELRA)},
	isbn = {978-2-9517408-7-7},
	language = {english}
}

@incollection{VaswaniNIPS2017_7181,
title = {Attention is All you Need},
author = {Vaswani, Ashish and Shazeer, Noam and Parmar, Niki and Uszkoreit, Jakob and Jones, Llion and Gomez, Aidan  and Kaiser, Lukasz and Polosukhin, Illia},
booktitle = {Advances in Neural Information Processing Systems 30},
pages = {5998--6008},
year = {2017},
publisher = {Curran Associates, Inc.},
url = {http://papers.nips.cc/paper/7181-attention-is-all-you-need.pdf}
}

@InProceedings{tensor2tensor-W18-1819,
  author = 	"Vaswani, Ashish
		and Bengio, Samy
		and Brevdo, Eugene
		and Chollet, Francois
		and Gomez, Aidan
		and Gouws, Stephan
		and Jones, Llion
		and Kaiser, {\L}ukasz
		and Kalchbrenner, Nal
		and Parmar, Niki
		and Sepassi, Ryan
		and Shazeer, Noam
		and Uszkoreit, Jakob",
  title = 	"{Tensor2Tensor} for Neural Machine Translation",
  booktitle = 	"Proceedings of the 13th Conference of the Association for Machine Translation in the Americas",
  year = 	"2018",
  publisher = 	"Association for Machine Translation in the Americas",
  pages = 	"193--199",
  url = 	"http://aclweb.org/anthology/W18-1819"
}

@inproceedings{LibriSpeech,
  author    = {Vassil Panayotov and
               Guoguo Chen and
               Daniel Povey and
               Sanjeev Khudanpur},
  title     = {{LibriSpeech}: An {ASR} corpus based on public domain audio books},
  booktitle = {{ICASSP}},
  pages     = {5206--5210},
  publisher = {{IEEE}},
  year      = {2015}
}

@inproceedings{DBLP:conf/icassp/ShenPWSJYCZWRSA18,
  author    = {Jonathan Shen and
               Ruoming Pang and
               Ron J. Weiss and
               Mike Schuster and
               Navdeep Jaitly and
               Zongheng Yang and
               Zhifeng Chen and
               Yu Zhang and
               Yuxuan Wang and
               RJ{-}Skerrv Ryan and
               Rif A. Saurous and
               Yannis Agiomyrgiannakis and
               Yonghui Wu},
  title     = {Natural {TTS} Synthesis by Conditioning Wavenet on {MEL} Spectrogram
               Predictions},
  booktitle = {{ICASSP}},
  pages     = {4779--4783},
  publisher = {{IEEE}},
  year      = {2018}
}

@inproceedings{Wang2017,
  author={Yuxuan Wang and R.J. Skerry-Ryan and Daisy Stanton and Yonghui Wu and Ron J. Weiss and Navdeep Jaitly and Zongheng Yang and Ying Xiao and Zhifeng Chen and Samy Bengio and Quoc Le and Yannis Agiomyrgiannakis and Rob Clark and Rif A. Saurous},
  title={{Tacotron: Towards End-to-End Speech Synthesis}},
  year=2017,
  booktitle={Proc. Interspeech},
  pages={4006--4010},
  doi={10.21437/Interspeech.2017-1452},
  url={http://dx.doi.org/10.21437/Interspeech.2017-1452}
}

@inproceedings{espnet,
  author={Shinji Watanabe and Takaaki Hori and Shigeki Karita and Tomoki Hayashi and Jiro Nishitoba and Yuya Unno and Nelson {Enrique Yalta Soplin} and Jahn Heymann and Matthew Wiesner and Nanxin Chen and Adithya Renduchintala and Tsubasa Ochiai},
  title={{ESPnet}: End-to-End Speech Processing Toolkit},
  year=2018,
  booktitle={Proc. Interspeech},
  pages={2207--2211},
  doi={10.21437/Interspeech.2018-1456},
  url={http://dx.doi.org/10.21437/Interspeech.2018-1456}
}

@article{Mikolov2010,
	author = {Mikolov, T and Karafiat, M and Burget, L and Cernocky, J and Khudanpur, S},
	journal = {Proc. Interspeech},
	pages = {1045--1048},
	title = {Recurrent Neural Network based Language Model},
	year = {2010}
}

@inproceedings{hori2018end,
  author    = {Takaaki Hori and
               Jaejin Cho and
               Shinji Watanabe},
  title     = {End-to-end Speech Recognition With Word-Based Rnn Language Models},
  booktitle = {2018 {IEEE} Spoken Language Technology Workshop, {SLT} 2018, Athens,
               Greece, December 18-21, 2018},
  pages     = {389--396},
  year      = {2018},
  url       = {https://doi.org/10.1109/SLT.2018.8639693},
  doi       = {10.1109/SLT.2018.8639693},
  bibsource = {dblp computer science bibliography, https://dblp.org}
}

@article{Bahdanau15,
	author    = {Dzmitry Bahdanau and
	Kyunghyun Cho and
	Yoshua Bengio},
	title     = {Neural Machine Translation by Jointly Learning to Align and Translate},
	journal   = {International Conference on Learning Representations},
	year      = {2015},
	url       = {http://arxiv.org/abs/1409.0473}
}

@inproceedings{post2013improved,
  Title = {Improved Speech-to-Text Translation with the {F}isher and {C}allhome {S}panish--{E}nglish Speech Translation Corpus},
  Author = {Post, Matt and Kumar, Gaurav and Lopez, Adam and Karakos, Damianos and Callison-Burch, Chris and Khudanpur, Sanjeev},
  Booktitle = {Proceedings of the International Workshop on Spoken Language Translation (IWSLT)},
  Year = {2013},
  Address = {Heidelberg, Germany},
  Month = {December}
}

@INPROCEEDINGS{kaldi-pitch,
author={P. {Ghahremani} and B. {BabaAli} and D. {Povey} and K. {Riedhammer} and J. {Trmal} and S. {Khudanpur}},
booktitle={ICASSP},
title={A pitch extraction algorithm tuned for automatic speech recognition},
year={2014},
volume={},
number={},
pages={2494-2498},
doi={10.1109/ICASSP.2014.6854049},
ISSN={1520-6149},}

@inproceedings{Zeyer2018,
  author={Albert Zeyer and Kazuki Irie and Ralf Schluter and Hermann Ney},
  title={Improved Training of End-to-end Attention Models for Speech Recognition},
  year=2018,
  booktitle={Proc. Interspeech},
  pages={7--11}
}

@inproceedings{Zhou2018,
  author={Shiyu Zhou and Linhao Dong and Shuang Xu and Bo Xu},
  title={Syllable-Based Sequence-to-Sequence Speech Recognition with the Transformer in {Mandarin Chinese}},
  year=2018,
  booktitle={Proc. Interspeech},
  pages={791--795},
  doi={10.21437/Interspeech.2018-1107},
  url={http://dx.doi.org/10.21437/Interspeech.2018-1107}
}

@article{Chan2016,
    author = {Chan, William and Jaitly, Navdeep and Le, Quoc and Vinyals, Oriol},
    doi = {10.1109/ICASSP.2016.7472621},
    isbn = {9781479999880},
    issn = {15206149},
    journal = {ICASSP},
    pages = {4960--4964},
    title = {{Listen, attend and spell: A neural network for large vocabulary conversational speech recognition}},
    volume = {2016-May},
    year = {2016}
}

@inproceedings{lakew-etal-2018-comparison,
    title = "A Comparison of Transformer and Recurrent Neural Networks on Multilingual Neural Machine Translation",
    author = "Lakew, Surafel Melaku  and
      Cettolo, Mauro  and
      Federico, Marcello",
    booktitle = "Proceedings of the 27th International Conference on Computational Linguistics",
    month = aug,
    year = "2018",
    address = "Santa Fe, New Mexico, USA",
    publisher = "Association for Computational Linguistics",
    url = "https://www.aclweb.org/anthology/C18-1054",
    pages = "641--652",
}

@inproceedings{ott-etal-2018-scaling,
    title = "Scaling Neural Machine Translation",
    author = "Ott, Myle  and
      Edunov, Sergey  and
      Grangier, David  and
      Auli, Michael",
    booktitle = "Proceedings of the Third Conference on Machine Translation: Research Papers",
    month = oct,
    year = "2018",
    address = "Belgium, Brussels",
    publisher = "Association for Computational Linguistics",
    url = "https://www.aclweb.org/anthology/W18-6301",
    pages = "1--9",
}

@INPROCEEDINGS{
         kaldi,
         author = {Povey, Daniel and Ghoshal, Arnab and Boulianne, Gilles and Burget, Lukas and Glembek, Ondrej and Goel, Nagendra and Hannemann, Mirko and Motlicek, Petr and Qian, Yanmin and Schwarz, Petr and Silovsky, Jan and Stemmer, Georg and Vesely, Karel},
          month = dec,
          title = {The {Kaldi} Speech Recognition Toolkit},
      booktitle = {IEEE 2011 Workshop on Automatic Speech Recognition and Understanding},
           year = {2011},
      publisher = {IEEE Signal Processing Society},
}

@article{luscher2019rwth,
title={{RWTH ASR} Systems for {LibriSpeech}: Hybrid vs Attention-w/o Data Augmentation},
  author={L{\"u}scher, Christoph and Beck, Eugen and Irie, Kazuki and Kitza, Markus and Michel, Wilfried and Zeyer, Albert and Schl{\"u}ter, Ralf and Ney, Hermann},
  journal={arXiv preprint arXiv:1905.03072},
  year={2019}
}

@article{irie2019language,
  title={Language Modeling with Deep Transformers},
  author={Irie, Kazuki and Zeyer, Albert and Schl{\"u}ter, Ralf and Ney, Hermann},
  journal={arXiv preprint arXiv:1905.04226},
  year={2019}
}

@inproceedings{HoriWZC17,
  author    = {Takaaki Hori and
               Shinji Watanabe and
               Yu Zhang and
               William Chan},
  title     = {Advances in Joint {CTC}-Attention Based End-to-End Speech Recognition
               with a Deep {CNN} Encoder and {RNN-LM}},
  booktitle = {Proc. Interspeech},
  pages     = {949--953},
  year      = {2017},
  url       = {http://www.isca-speech.org/archive/Interspeech\_2017/abstracts/1296.html},
  bibsource = {dblp computer science bibliography, https://dblp.org}
}

@INPROCEEDINGS{speech-transformer,
author={L. {Dong} and S. {Xu} and B. {Xu}},
booktitle={ICASSP},
title={Speech-Transformer: A No-Recurrence Sequence-to-Sequence Model for Speech Recognition},
year={2018},
volume={},
number={},
pages={5884-5888},
doi={10.1109/ICASSP.2018.8462506},
ISSN={2379-190X},}

@inproceedings{CrossVila2018,
  author={Laura {Cross Vila} and Carlos Escolano and José A. R. Fonollosa and Marta {R. Costa-Jussà}},
  title={End-to-End Speech Translation with the Transformer},
  year=2018,
  booktitle={Proc. IberSPEECH 2018},
  pages={60--63},
  doi={10.21437/IberSPEECH.2018-13},
  url={http://dx.doi.org/10.21437/IberSPEECH.2018-13}
}

@inproceedings{li2019close,
  title={Neural Speech Synthesis with Transformer Network},
  author={Naihan Li and Shujie Liu and Yanqing Liu and Sheng Zhao and Ming Liu and Ming Tou Zhou},
  booktitle={The AAAI Conference on Artificial Intelligence (AAAI)},
  year={2019}
}

@inproceedings{tachibana2018efficiently,
  title={Efficiently trainable text-to-speech system based on deep convolutional networks with guided attention},
  author={Tachibana, Hideyuki and Uenoyama, Katsuya and Aihara, Shunsuke},
  booktitle={ICASSP},
  pages={4784--4788},
  year={2018},
  organization={IEEE}
}

@ARTICLE{fastspeech,
       author = {{Ren}, Yi and {Ruan}, Yangjun and {Tan}, Xu and {Qin}, Tao and
         {Zhao}, Sheng and {Zhao}, Zhou and {Liu}, Tie-Yan},
        title = "{FastSpeech}: Fast, Robust and Controllable Text to Speech",
      journal = {arXiv e-prints},
         year = "2019",
        month = "May",
          eid = {arXiv:1905.09263},
        pages = {arXiv:1905.09263},
archivePrefix = {arXiv},
       eprint = {1905.09263},
 primaryClass = {cs.CL},
       adsurl = {https://ui.adsabs.harvard.edu/abs/2019arXiv190509263R}
}

@inproceedings{Weiss2017,
  author={Ron J. Weiss and Jan Chorowski and Navdeep Jaitly and Yonghui Wu and Zhifeng Chen},
  title={Sequence-to-Sequence Models Can Directly Translate Foreign Speech},
  year=2017,
  booktitle={Proc. Interspeech},
  pages={2625--2629},
  doi={10.21437/Interspeech.2017-503},
  url={http://dx.doi.org/10.21437/Interspeech.2017-503}
}


\end{document}